\documentclass[letterpaper]{article} 
\usepackage{aaai24}  
\usepackage{times}  
\usepackage{helvet}  
\usepackage{courier}  
\usepackage[hyphens]{url}  
\usepackage{graphicx} 
\usepackage{natbib}  
\usepackage{caption} 
\usepackage{algorithm}
\usepackage{algorithmic}
\usepackage{graphicx}
\usepackage{amsmath}
\usepackage{amssymb}
\usepackage{booktabs}
\usepackage{times}
\usepackage{helvet}
\usepackage{courier}
\usepackage{epsfig}
\usepackage{mathrsfs}
\usepackage{multirow}
\usepackage{color}
\usepackage{xcolor}
\usepackage{colortbl}
\usepackage{tabularx}
\usepackage{arydshln}
\usepackage{pifont}

\usepackage{newfloat}
\usepackage{listings}
\DeclareCaptionStyle{ruled}{labelfont=normalfont,labelsep=colon,strut=off} 
\floatstyle{ruled}
\newfloat{listing}{tb}{lst}{}
\floatname{listing}{Listing}
\title{MatchDet: A Collaborative Framework for Image Matching and Object Detection}
\author {
    Jinxiang Lai\equalcontrib\textsuperscript{\rm 1},
    Wenlong Wu\equalcontrib\textsuperscript{\rm 1},
    Bin-Bin Gao\textsuperscript{\dag\rm 1},
    Jun Liu\textsuperscript{\rm 1},
    Jiawei Zhan\textsuperscript{\rm 1},
    Congchong Nie\textsuperscript{\rm 1},
    Yi Zeng\textsuperscript{\rm 1},
    Chengjie Wang\thanks{Corresponding Author}\textsuperscript{\rm 1, 2}
}
\affiliations {
    \textsuperscript{\rm 1} Tencent Youtu Lab, China \\
    \textsuperscript{\rm 2} Shanghai Jiao Tong University, China\\
    layjins1994@gmail.com, ezrealwu@tencent.com, \{csgaobb, junsenselee\}@gmail.com,
    \{gavynzhan, nickccnie, sylviazeng\}@tencent.com, jasoncjwang@tencent.com
}

\usepackage{bibentry}

\begin{document}

\maketitle

\begin{figure}[t!]
\centering
\includegraphics[width=0.99\linewidth]{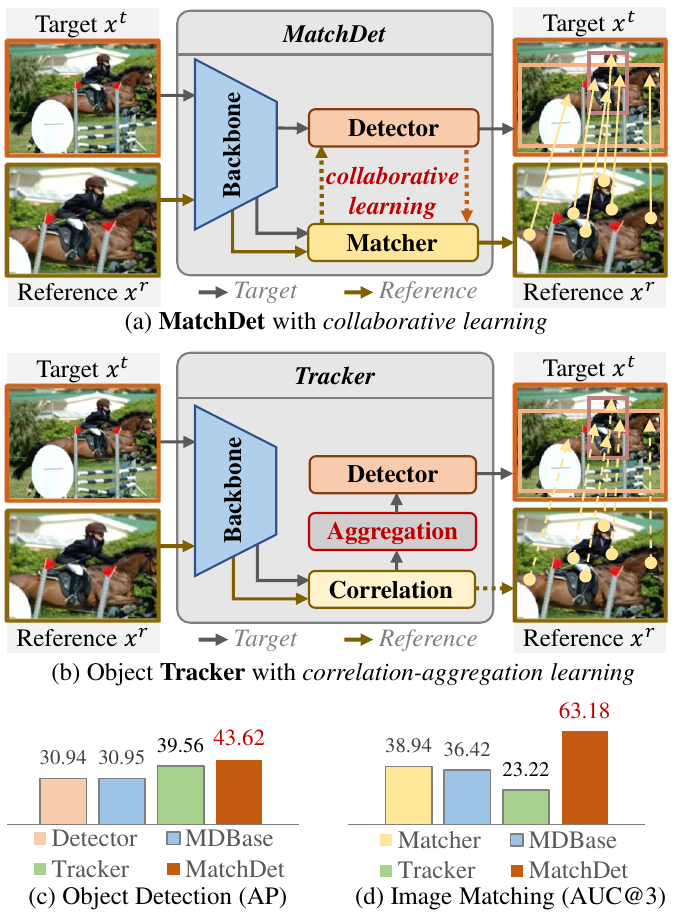}
\caption{(a) Our MatchDet with collaborative learning for improving image matching and object detection. We introduce a baseline named MDBase network, which removes the collaborative learning module of MatchDet. (b) The object Tracker with correlation-aggregation learning. The dashed line represents that the Tracker has the potential ability to obtain pairwise correspondences, while there is no matching objective function to supervise it. (c) and (d) are the results on Warp-COCO dataset. (c) Our MatchDet obtains 4.06\% improvement in object detection. (d) Our MatchDet achieves 24.24\% higher performance in image matching.}
\label{fig:MatchDet_intro}
\end{figure}

\begin{abstract}
Image matching and object detection are two fundamental and challenging tasks, while many related applications consider them two individual tasks (i.e. task-individual).
In this paper, a collaborative framework called MatchDet (i.e. task-collaborative) is proposed for image matching and object detection to obtain mutual improvements.
To achieve the collaborative learning of the two tasks, we propose three novel modules, including a Weighted Spatial Attention Module (WSAM) for Detector, and Weighted Attention Module (WAM) and Box Filter for Matcher.
Specifically, the WSAM highlights the foreground regions of target image to benefit the subsequent detector, the WAM enhances the connection between the foreground regions of pair images to ensure high-quality matches, and Box Filter mitigates the impact of false matches.
We evaluate the approaches on a new benchmark with two datasets called Warp-COCO and miniScanNet.
Experimental results show our approaches are effective and achieve competitive improvements.
\end{abstract}

\section{Introduction}
\label{sec:intro}
Image matching \cite{shrivastava2011data} and object detection \cite{liu2020deep} are two fundamental and challenging tasks in computer vision.
Image matching finds pixel-wise correspondences between image pairs, and object detection seeks to locate and classify object instances in images.
With the combination of them, there are numerous important applications, including robot vision, autonomous driving, and industrial defect inspection.
In robot vision and autonomous driving, it usually uses image matching technique to perform Simultaneous Localization And Mapping (SLAM) \cite{mur2015orb}, and also needs to find the target category objects (e.g. pedestrian in autonomous driving) in images based on object detection technique.
In industrial defect inspection, it applies image matching for registration \cite{shrivastava2011data} to obtain Region-of-Interest (ROI), and then detects the target defects.
The aforementioned applications consider image matching and object detection as two individual tasks (i.e. task-individual).

In this paper, a collaborative framework called MatchDet (i.e. task-collaborative) is proposed for image matching and object detection to obtain mutual improvements.
As illustrated in Fig.\ref{fig:MatchDet_intro}(a), given input reference and target images, our MatchDet simultaneously outputs their homography relationship and object detection results of the target image, which is defined as a Match-and-Detection task.
The proposed MatchDet framework consists of a shared backbone, and two Matcher and Detector task branches for image matching and object detection, which is co-trained end-to-end.
Under task-collaborative MatchDet framework, the homography relationship estimated by Matcher and the bounding boxes predicted by Detector can be useful to each other.

As presented in Fig.\ref{fig:MatchDet_intro}(b), the most relevant approach is the correlation based Tracker \cite{densetracker} for the object tracking task.
The Tracker utilizes the Correlation module to explore the affinity between the target image and the reference image, then further applies the Aggregation module to refine the affinity for enhancing the target objects.
There are two main differences between MatchDet and Tracker:
(i) MatchDet integrates a Matcher branch, which is able to obtain a precise homography relationship. However, Tracker only adopts the Correlation module to implicitly explore the affinity without the supervision of matching objective function, which leads to an imprecise homography relationship. As illustrated in Fig.\ref{fig:MatchDet_intro}(d), our MatchDet achieves a large improvement with 39.96\% in image matching task.
(ii) MatchDet achieves mutual performance improvements in the two tasks via the proposed collaborative learning module, while Tracker only utilizes correlation-aggregation learning to improve the performance of the object detection task.

To achieve mutual performance improvements via the collaborative learning of the two tasks, we propose three novel modules, including a Weighted Spatial Attention Module (WSAM) for Detector, Weighted Attention Module (WAM) and Box Filter for Matcher.
For Matcher branch, the proposed WAM produces more discriminative feature representations of image pairs, via learning global context to find the correspondences among surrounding regions, with the usage of Transformer structure \cite{sun2021loftr}.
Benefiting from the Match-and-Detection task, it's achievable to obtain the potential foreground regions of images.
Then the WAM can be more focus on interacting information among the foreground regions of pairs, which reduces background interference and ensures high-quality matches.
Specifically, the WAM first utilizes a Weights Generator to generate weighted maps based on the foreground regions, and then uses the weighted maps to enhance the affinity matrix between feature pairs as implementing in the Attention operation of Transformer.
Further more, Box Filter reduces the impact of the potential low-quality matches, via strengthening the matching scores among the foreground regions of pairs, where the foregrounds are predicted by Detector.

For Detector branch, the proposed WSAM, a variant of WAM, highlights the foreground regions of target image, via the similar regions with instance feature of reference image and learnable semantic embedding.
To achieve the above purpose of enhancing foregrounds, the WSAM adopts Weighted Spatial Attention instead of Weighted Attention applied in WAM, and their essential difference is discussed in APPENDIX.
Similar to WAM, the WSAM also pays more attention on feature interaction among foregrounds.
The WSAM transforms the foreground regions of reference image with the homography estimated by Matcher, to generate the potential foreground regions for target image.

In general, our main contributions are:

$\bullet$ For the first time, we propose a collaborative framework called MatchDet for image matching and object detection to obtain mutual improvements. Besides, our MatchDet is a general framework, which can utilize different detectors such as FCOS, Faster-RCNN, and AdaMixer.

$\bullet$ To achieve collaborative learning of image matching and object detection, three novel modules are proposed, including a Weighted Spatial Attention Module (WSAM) which highlights the foreground regions of target image for Detector, and Weighted Attention Module (WAM) and Box Filter which obtains high-quality matches for Matcher.

$\bullet$ We evaluate the Match-and-Detection task on a new benchmark with two datasets called Warp-COCO and miniScanNet.
This benchmark can be used to verify the performances of the algorithms on both image matching and object detection.
Experimental results show our approaches are effective and achieve competitive improvements.

\begin{figure*}[!t]
\centering
\includegraphics[width=0.99\linewidth]{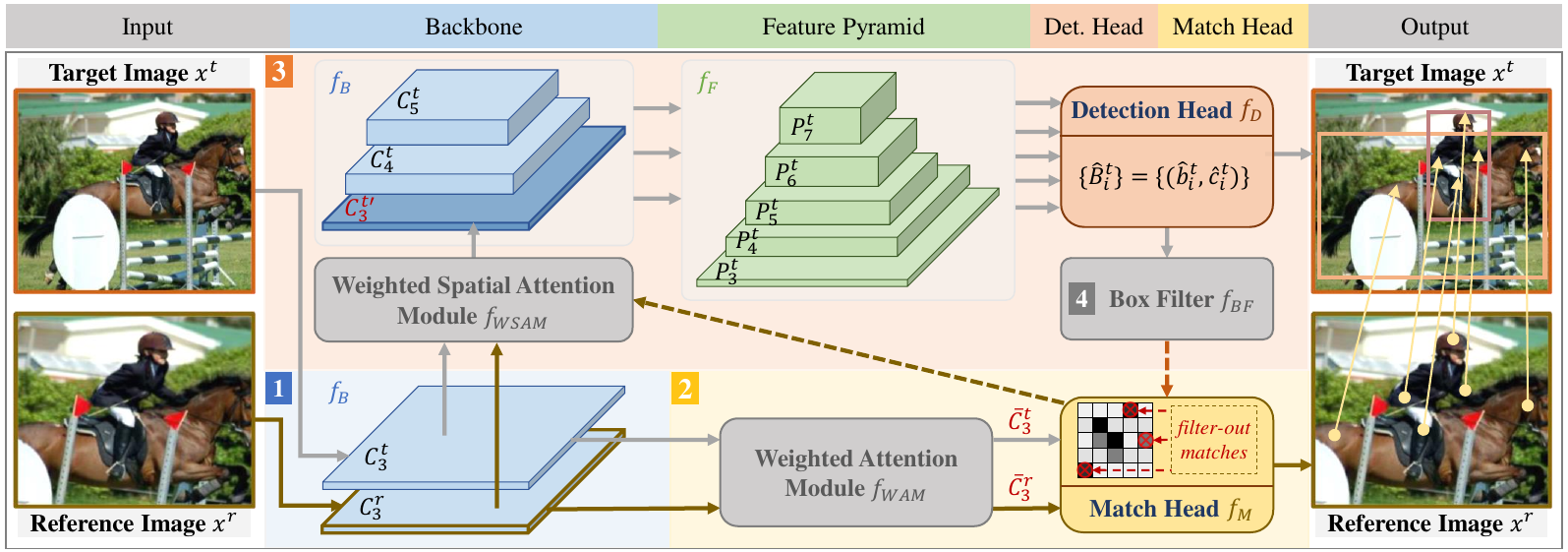}
\caption{The network architecture of our MatchDet.
There are four stages: \ding{172} Obtaining basic features $\{{C^t_3},{C^r_3}\}$ with a shared backbone.
\ding{173} Matcher branch estimates the homography matrix with the enhanced features $\{\bar{C}^t_3,\bar{C}^r_3\}$ produced by Weighted Attention Module.
\ding{174} Detector branch predicts the bounding boxes based on the highlighted features ${C^t_3}'$ generated by Weighted Spatial Attention Module.
\ding{175} Box Filter refines the image matching results via filtering out the potential mismatches.
}
\label{fig:MatchDet}
\end{figure*}

\section{Related Work}
\noindent\textbf{Object Detection} \
The current object detection algorithms can be divided into anchor-based \cite{fasterRCNN,YOLOv3,retinaNet,ATSS}, anchor-free \cite{cornerNet,centerNet,tian2019fcos}, and query-based \cite{carion2020end,gao2022adamixer,zhang2022dino} methods.
In this paper, we choose the classical anchor-free FCOS \cite{tian2019fcos} as the basic detector due to its good performance and simplicity.

\noindent\textbf{Image Matching} \
Image matching finds pixel-wise correspondences between image pairs, with two main directions of Interest Point Detector-based (IPD-based) methods \cite{rublee2011orb,detone2018superpoint} and IPD-free methods \cite{sun2021loftr,chen2022guide,huang2023adaptive}.
In this paper, we choose the classical IPD-free LoFTR \cite{sun2021loftr} as the basic image matcher due to its good performance and simplicity.

\noindent\textbf{Transformer Attention}
The transformer Cross Attention in LoFTR \cite{sun2021loftr} learns global context based on the full affinity matrix between feature pairs, but it may easily be distracted by background and causes the slow convergence of model.
To alleviate this problem, we propose two novel transformer-based weighted attention and weighted spatial attention modules to strengthen the target object.


\section{Problem Definition}
\label{sec:problem_def}
The Match-and-Detection task aims to obtain the homography relationship and object detection results of the input pair images.
Formally, given the pair images ${x^t}$ and ${x^r}$, the Match-and-Detection method predicts the homography matrix $\hat{\mathcal{H}}$ and bounding boxes $\{\hat{B}_i^t\}$ for ${x^t}$.
We denote ${x^t}$ and ${x^r}$ as the target and reference images respectively, the homography matrix $\mathcal{H}\in \mathbb{R}^{3 \times 3}$ represents their geometric relationship as ${x^t}=\mathcal{H}{x^r}$, and their ground-truth bounding boxes are defined as $\{B_i^t\}$ and $\{B_i^r\}$ respectively.
Here $B_i = (b_{i}, c_{i}) \in \mathbb{R}^{4} \times \{1, 2, \cdots, C\}$, $b_{i}$ and $c_{i}$ denote the bounding box and the category of the object respectively, and $C$ is the classes number of dataset.

In training stage, all ground-truth labels are given.
In inference stage, with different sources of bounding boxes of reference image, we introduce three different settings:
(a) \textbf{GTBoxR} setting, gives Ground-Truth bounding Boxes $\{B_i^r\}$ of Reference image in inference. Some of the corresponding applications are robotic Pick-and-Place \cite{zeng2018robotic} and robot navigation for known scenarios, of which the reference image can be stored and labeled in advance.
(b) \textbf{PreBoxR} setting, gives Prediction bounding Boxes $\{\hat{B}_i^r\}$ of Reference image in inference. It is suitable for the video-based scenarios, which can reuse the prediction results of the previous frame image.
(c) \textbf{NoBoxR} setting,  gives No bounding Boxes of Reference image in inference. It is more general and challenging than the above settings.

\section{Methodology}
We first propose a simple baseline method called MDBase (Match-and-Detection Baseline) network ${f_{base}}$, which adopts multi-task paradigm with a shared backbone ${f_B}$ and two task-heads (i.e. Match Head ${f_M}$ and Detection Head ${f_D}$) for matching and detection.
Then, as illustrated in Fig.\ref{fig:MatchDet}, the MatchDet network ${f_{MD}}$ is constructed upon MDBase network, which additionally inserts three novel modules including Weighted Spatial Attention Module (WSAM) ${f_{WSAM}}$, Weighted Attention Module (WAM) ${f_{WAM}}$ and Box Filter ${f_{BF}}$.
The WSAM highlights the foreground regions of target image, the WAM enhances the connection between the foreground regions of pair images, and Box Filter mitigates the impact of false matches.

Specifically, let ${f^i_B}$ be the ${i^{th}}$ layer of backbone ${f_B}$, ${f^i_F}$ be the ${i^{th}}$ layer of FPN (Feature Pyramid Network) ${f_F}$.
Given input image pairs $\{{x^t},{x^r}\}$, the backbone ${f_B}$ generates corresponding target features $\{C^t_3, C^t_4, C^t_5\}$ and reference features $C^r_3$ respectively, and the FPN ${f_F}$ further produces target features $\{P^t_3, P^t_4, P^t_5, P^t_6, P^t_7\}$.
In particular, the ${i^{th}}$ layer features $C^t_i$ and $C^r_i$ are produced by ${f^i_B}$, and the same goes for other features like $P^t_i$ by ${f^i_F}$, where $\{C^t_i, C^r_i, P^t_i\} \in \mathbb{R}^{c_i\times h_i\times w_i}$.
Then, WSAM and WAM generate the enhanced features ${C^t_3}'$ and $\{\bar{C}^t_3,\bar{C}^r_3\}$ for Detector and Matcher, respectively.
Finally, the Match Head ${f_M}$ and Detection Head ${f_D}$ predict the homography matrix $\hat{\mathcal{H}}$ and bounding boxes $\{\hat{B}_i^t\}$ for ${x^t}$.
In detail, the Detection Head ${f_D}$ is the same as FCOS, and the Match Head ${f_M}$ is similar to LoFTR which utilizes a dual-softmax matching layer.

\subsection{MDBase Network}
The MDBase network consists of a shared backbone and two task-heads for matching and detection, i.e. ${f_{base}} = [{f_B},{f_F},{f_M},{f_D}]$.
The MDBase makes predictions as follows:
\begin{equation}
\begin{aligned}
\{\hat{B}_i^t\}, \hat{\mathcal{H}} &= {f_{base}}\left(x^t,x^r\right) \\
where, \, \{\hat{B}_i^t\} &= f_D\left(f_F\left(f_B\left(x^t\right)\right)\right), \\
\hat{\mathcal{H}} &= f_M\left(f_B\left(x^t,x^r\right)\right)
\label{eq:MDBase}
\end{aligned}
\end{equation}
Comparing to the current task-individual solutions, the MDBase network has lower complexity by utilizing a shared backbone.
In order to obtain mutual improvements of image matching and object detection, we next introduce a novel MatchDet network benefiting from the collaborative learning of the two tasks.

\subsection{MatchDet Network}
As illustrated in Fig.\ref{fig:MatchDet}, on the basic of MDBase network, the proposed MatchDet integrates Weighted Spatial Attention Module ${f_{WSAM}}$, Weighted Attention Module ${f_{WAM}}$ and Box Filter ${f_{BF}}$ to achieve collaborative learning of image matching and object detection tasks.
Formally, MatchDet consists of ${f_{MD}} = [f_{base},{f_{WSAM}},{f_{WAM}},{f_{BF}}]$, and makes predictions as:
\begin{equation}
\begin{aligned}
\{\hat{B}_i^t\}, \hat{\mathcal{H}} &= {f_{MD}}\left(x^t,x^r\right) \\
where, \, \{\hat{B}_i^t\} &= f_D\left(f_F\left({f_{WSAM}}{(}f_B(x^t,x^r){)}\right)\right), \\
 \hat{\mathcal{H}} &= f_M\left({f_{BF}}{(}{f_{WAM}}{(}f_B(x^t,x^r){)}{)}\right)
\label{eq:MatchDet}
\end{aligned}
\end{equation}
In the following, we introduce the detailed structure of the proposed modules and loss function.

\begin{figure}[t!]
\centering
\includegraphics[width=0.95\linewidth]{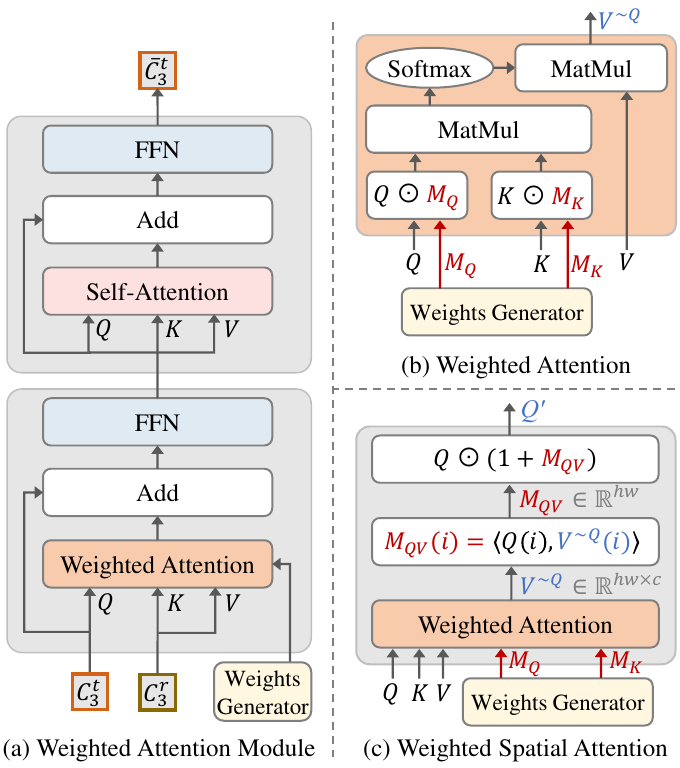}
\caption{(a) The Weighted Attention Module (WAM) consists of a Weighted Attention block and a Self-Attention block, where $\{Q,K,V\}$ are known as $\{\emph{query}, \emph{key}, \emph{value}\}$ and FFN denotes Feed-Forward Network in Transformer.
(b) The Weighted Attention applied in WAM, where ${\odot}$ is Broadcasting Element-wise Product. The variables dimensions are ${\{V^{\sim Q},Q,K,V\} \in \mathbb{R}^{hw \times c }}$ and ${\{M_Q,M_K\} \in \mathbb{R}^{hw}}$.
(c) The Weighted Spatial Attention enhances the spatial response of $Q$ by ${M_{QV} \in \mathbb{R}^{hw}}$ to obtain ${Q' \in \mathbb{R}^{hw \times c }}$, where $\langle \cdot \rangle$ calculates the cosine similarity. And replacing Weighted Attention of WAM with Weighted Spatial Attention derives the Weighted Spatial Attention Module (WSAM).}
\label{fig:WAM}
\end{figure}

\begin{figure}[t!]
\centering
\includegraphics[width=0.90\linewidth]{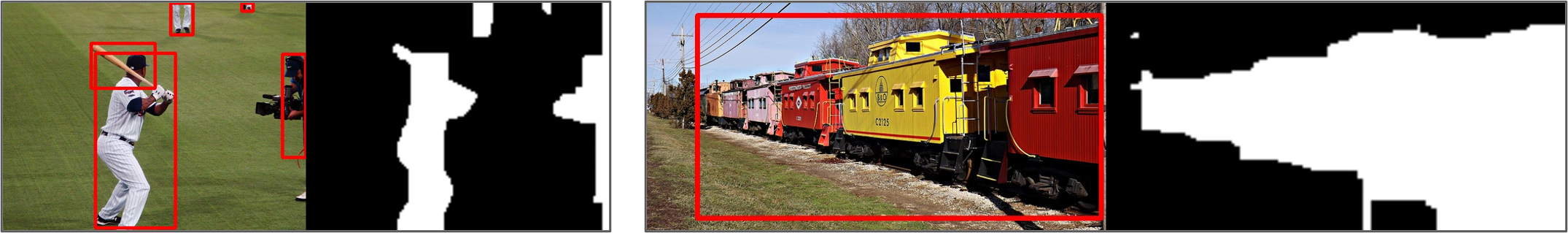}
\caption{The visualizations of the generated Weighted Map.}
\label{fig:mask}
\end{figure}

\subsubsection{Weighted Attention Module}
The Weighted Attention Module ${f_{WAM}}$ enhances the connection between the foreground regions of pair images, which is expressed as:
\begin{equation}
\begin{aligned}
\bar{C}^t_3,\bar{C}^r_3 = f_{WAM}\left(C^t_3,C^r_3\right),f_{WAM}\left(C^r_3,C^t_3\right)
\label{eq:WAM}
\end{aligned}
\end{equation}
As shown in Fig.\ref{fig:WAM}(a), the WAM stacks a Weighted Attention block and a Self-Attention block.
Different from the Cross Attention used in LoFTR \cite{sun2021loftr} and Masked Attention applied in Mask2Former \cite{cheng2022masked}, we propose a novel Weighted Attention as presented in Fig.\ref{fig:WAM}(b), which enhances $\{Q,K\}$ with two different weighted maps $\{M_Q,M_K\}$ produced by Weights Generator, respectively.
The Weighted Attention operation is formulated as:
\begin{equation}
\begin{aligned}
V^{\sim Q}&=\operatorname{WeightedAttention}\left(\left(Q,K,V\right),M_Q,M_K\right)\\
&=\sigma\left(\left(Q \odot M_Q\right)\left(K \odot M_K\right)^T\right)V
\label{eq:WA}
\end{aligned}
\end{equation}
where ${\sigma}$ is softmax function. The weighted maps ${\{M_Q,M_K\} \in \mathbb{R}^{hw}}$ represent the foreground regions of pairs $\{Q,K\}$.
Under three different settings introduced in Problem Definition, the Weights Generator of WAM produces different weighted maps $\{M_Q,M_K\}$ as follows:

$\bullet$ \textbf{GTBoxR} setting. We denote ${\{M_t,M_r\} \in \mathbb{R}^{hw}}$ as the foreground regions of $\{C^t_3,C^r_3\}$.
In the case of implementing $\bar{C}^t_3 = f_{WAM}\left(C^t_3,C^r_3\right)$, there are $M_Q=M_t$ and $M_K=M_r$.
Calculating $\bar{C}^r_3 = f_{WAM}\left(C^r_3,C^t_3\right)$ assigns $M_Q=M_r$ and $M_K=M_t$. The following assignment rules of $\{M_Q,M_K\}$ with $\{M_t,M_r\}$ are the same, thus we only introduce how to generate $\{M_t,M_r\}$ for simplicity.
(a) This setting gives ground-truth $\{B_i^r\}$ for reference image, then we generate $M_r$ by assigning the $\{B_i^r\}$ regions as $1+{\alpha}_1$ and other background regions as $1$.
(b) For target feature ${C}^t_3$, we first use a light decoder to predict a semantic segmentation mask $m_t$. Then we produce $M_t$ by assigning the foreground regions of $m_t$ that are higher than $0.5$ as $1+{\alpha}_1$, and other background regions as $1$.
The detail of the light segmentation decoder is described in the following.

$\bullet$ \textbf{PreBoxR} setting. Similarly, we use the predicted $\{\hat{B}_i^r\}$ and $m_t$ to generate $M_r$ and $M_t$, respectively.

$\bullet$ \textbf{NoBoxR} setting. The light decoder predicts two semantic segmentation mask $m_r$ and $m_t$ to generate $M_r$ and $M_t$.

The light segmentation decoder consists of two $3 \times 3$ convolution layers and a $1 \times 1$ convolution layer, which per-pixel classifies the foreground and background of the image.
The light segmentation decoder is optimized by the box-supervised segmentation loss proposed in BoxInst \cite{tian2021boxinst}, which utilizes the ground-truth bounding boxes to supervise the segmentation decoder.
Fig.\ref{fig:mask} presents the generated Weighted Map as the foreground regions.

There are two important abilities of WAM:
(a) The features $\{\bar{C}^t_3,\bar{C}^r_3\}$ processed through WAM are more discriminative than $\{{C}^t_3,{C}^r_3\}$.
The dense local features $\{{C}^t_3,{C}^r_3\}$ extracted by Convolutional Neural Networks (CNNs) have limited receptive field which may not distinguish indistinctive or repetitive regions.
Transforming by WAM, the global features $\{\bar{C}^t_3,\bar{C}^r_3\}$ have larger global context to find the correspondences among surrounding regions.
(b) The WAM enhances the connection between foreground regions with the constrained of $\{M_Q,M_K\}$, which benefits the subsequent matching layer to obtain high-quality matches.
The Cross Attention in Transformer learns global context based on the affinity matrix between feature pairs, which may easily to be disturbed by background noise and leads to false matches.
With the constrained of $\{M_Q,M_K\}$, the WAM can be more focus on interacting information among the foreground regions to ensure high-quality matches.

\subsubsection{Weighted Spatial Attention Module}
The Weighted Spatial Attention Module ${f_{WSAM}}$ highlights the foreground regions of target image, which is expressed as:
\begin{equation}
\begin{aligned}
{C^t_3}' = f_{WSAM}\left(C^t_3,C^r_3\right)
\label{eq:WSAM}
\end{aligned}
\end{equation}
The WSAM stacks a Weighted Spatial Attention block and a Self-Attention block. And the structure of the key component Weighted Spatial Attention (WSAttention) is illustrated in Fig.\ref{fig:WAM}(c), which is formulated as:
\begin{equation}
\begin{aligned}
&Q'=\operatorname{WSAttention}\left(\left(Q,K,V\right),M_Q,M_K\right)\\
&\quad=Q \odot (1 + M_{QV}) \\
&where, \, M_{QV}(i) = \langle{Q(i), V^{\sim Q}(i)}\rangle, \\
&V^{\sim Q}=\operatorname{WeightedAttention}\left(\left(Q,K,V\right),M_Q,M_K\right)
\label{eq:WSA}
\end{aligned}
\end{equation}
Here we explain the meanings of key variables in Weighted Spatial Attention:
(a) The $V^{\sim Q}$, is $Q$-aligned $V$, i.e. aggregates $V$ into alignment with $Q$. At $i^{th}$ spatial position, $V^{\sim Q}(i)$ is aggregated from $V$ with the affinity vector between ${(Q(i),K)}$. Intuitively, $V^{\sim Q}(i)$ collects all the patch features from $V$ that are semantically similar to ${Q(i)}$.
(b) $Q{'}$, is $M_{QV}$-enhanced $Q$, i.e. uses the cosine similarity map $M_{QV}$ between $({Q},V^{\sim Q})$ to re-weight $Q$.
Therefore, ${Q{'}}$ will be enhanced if ${Q}$ is similar to $V^{\sim Q}$ (i.e. $Q$-aligned $V$).
Thus Weighted Spatial Attention is able to enhance the spatial response of $Q$ via the similar regions with $V$.

Based on the above analysis, we propose to use instance feature ${C^r_3}$ and learnable semantic embedding $W_e\in \mathbb{R}^{C\times c_3}$ (i.e. pytorch code is $W_e$=nn.Embedding($C$, $c_3$)) to highlight ${C^t_3}$, i.e. ${C^t_3}$ can be highlighted by the similar regions of ${C^r_3}$ and $W_e$, where $W_e$ contains foreground semantics via learning all categories information \cite{lai2022tsf}.
Formally,
\begin{equation}
\begin{aligned}
Q'=&\operatorname{WSAttention}\left(\left(C^t_3,C^r_3,C^r_3\right),M_Q,M_K\right)\\
&+ \operatorname{WSAttention}\left(C^t_3,W_e,W_e\right)
\label{eq:WSA2}
\end{aligned}
\end{equation}

The WSAM needs to find the foreground regions of ${C^r_3}$ to highlight ${C^t_3}$.
Under different settings, the Weights Generator of WSAM produces different weighted maps $\{M_Q,M_K\}$=$\{M_t,M_r\}$ (i.e. the potential foreground regions of $\{C^t_3,C^r_3\}$) as follows:

$\bullet$ \textbf{GTBoxR} setting.
For ${C}^r_3$, we generate $M_r$ by assigning the given ground-truth $\{B_i^r\}$ regions as $1+{\alpha}_2$ and other background regions as $1$.
For ${C}^t_3$, we use the estimated homography ${\mathcal{H}}'$ = $f_M\left({f_{WAM}}\left(f_B\left(x^t,x^r\right)\right)\right)$ to affine $M_r$ to generate $M_t$ = ${\mathcal{H}}'M_r$.

$\bullet$ \textbf{PreBoxR} setting. Similarly, we use the predicted $\{\hat{B}_i^r\}$ to generate $M_r$, and then obtains $M_t$ = ${\mathcal{H}}'M_r$.

$\bullet$ \textbf{NoBoxR} setting. Firstly, the light decoder predicts two semantic segmentation mask $m_r$ and $m_t$ to generate $M_r$ and $M_t$, respectively. Then we further refine the maps by $M_r$ = $M_r+{\mathcal{H}}'^{-1}M_t$ and $M_t$ = $M_t+{\mathcal{H}}'M_r$.

\subsubsection{Match Head with Box Filter}
\noindent\textbf{Box Filter $f_{BF}$} produces an filter map $\mathcal{F}$ to mitigate the impact of false matches measured by ${\mathcal{H}}'$ = $f_M\left({f_{WAM}}\left(f_B\left(x^t,x^r\right)\right)\right)$ on stage \ding{173}.
After stage \ding{174}, Detector branch outputs the prediction boxes $\{\hat{B}_i^t\}$.
On stage \ding{175}, we first generate foreground regions $\hat{M}_t$ of target image by assigning $\{\hat{B}_i^t\}$ regions as $1+\beta$ and other background regions as $1$, and then we obtain the foreground $\hat{M}_r$ of reference image from WSAM with ${M}_r$ of which foreground is also re-assigned as $1+\beta$.
Finally, filter map $\mathcal{F}$ is generated by
$\mathcal{F}(i, j) = \hat{M}_t(i) \cdot \hat{M}_r(j)$.

\noindent\textbf{Match Head $f_{M}$} is based on the dual-softmax matching layer \cite{tyszkiewicz2020disk,sun2021loftr}.
Formally, the matching probability $\mathcal{P}$ is obtained by:
\begin{equation}
\begin{aligned}
\mathcal{P}(i, j) = \sigma\left(\mathcal{S}\left(i, \cdot \right)\right)_j \cdot \sigma\left(\mathcal{S}\left(\cdot, j\right)\right)_i, \\
where, \, \mathcal{S}\left(i, j\right) = \frac{1}{\tau} \cdot \langle\bar{C}^t_3(i), \bar{C}^r_3(j)\rangle
\label{eq:dual-softmax}
\end{aligned}
\end{equation}
where $\mathcal{S}$ is the score matrix between the feature pairs $(\bar{C}^t_3, \bar{C}^r_3)$.
The dual-softmax matching applies softmax on both dimensions of $\mathcal{S}$ to obtain the matching probability $\mathcal{P}$.
Then, we use the filter map $\mathcal{F}$ generated by Box Filter to update $\mathcal{P}$ by
$\mathcal{P}(i, j) \leftarrow \mathcal{P}(i, j) \cdot \mathcal{F}(i, j)$.
Next, we obtain the match prediction by
$\hat{\mathcal{M}} = \{\left(\tilde{i},\tilde{j}\right) \mid \forall\left(\tilde{i},\tilde{j}\right) \in \operatorname{MNN}\left(\mathcal{P}\right),\ \mathcal{P}\left(\tilde{i},\tilde{j}\right) \geq \theta\}$,
where, MNN is a Mutual Nearest Neighbor operator, and $\theta$ is a threshold to select good matches.
Finally, the homography matrix is estimate by RANSAC algorithm with
$\hat{\mathcal{H}} = \operatorname{RANSAC}(\hat{\mathcal{M}})$.

\subsubsection{Loss Function}
The MatchDet network is optimized by the loss function:
$\mathcal{L} = \mathcal{L}_{matcher} + \lambda \mathcal{L}_{detector}$,
where $\mathcal{L}_{matcher}$ and $\mathcal{L}_{detector}$ are the losses of Matcher branch and Detector branch respectively, and $\lambda$ is the balance weight.
The $\mathcal{L}_{detector}$ is the same as FCOS \cite{tian2019fcos}.
Following LoFTR \cite{sun2021loftr}, the $\mathcal{L}_{matcher}$ is formulated as:
$\mathcal{L}_{matcher} = - \frac{1}{\lvert\mathcal{M}\rvert}
\sum_{(\tilde{i}, \tilde{j}) \in \mathcal{M}} \log \hat{\mathcal{M}}\left(\tilde{i}, \tilde{j}\right)$,
where $\mathcal{M}$ and $\hat{\mathcal{M}}$ are ground-truth and predicted matches.

\renewcommand{\tabcolsep}{3.5pt}
\begin{table*}[t!]
\centering
\small
\begin{tabular}{ l | l | c c c | c c c | c c c | c c}
\hline
\multicolumn{1}{l|}{\{Target image,} & \multicolumn{1}{l|}{\multirow{2}*{Method}} & \multicolumn{6}{c|}{\emph{Average Precision} on Target image} & \multicolumn{3}{c}{Homography est. AUC} & \multicolumn{1}{|c}{\multirow{2}*{Params}} & \multicolumn{1}{c}{\multirow{2}*{FLOPs}}\\
\cline{3-11} \multicolumn{1}{l|}{Reference image\}} & \multicolumn{1}{c|}{}  &AP & AP$_{50}$ & AP$_{75}$ & AP$_{S}$ & AP$_{M}$ & AP$_{L}$ &@3px &@5px &@10px & \multicolumn{1}{c}{} & \multicolumn{1}{c}{}\\
\hline
&FCOS \cite{tian2019fcos} &31.02 &47.99 &33.08 & 11.09 & 28.10 & 43.66 & - & - & - & 32.29M & 174604M \\
&LoFTR \cite{sun2021loftr}  & - & - & - & - & - & - & 47.94 & 68.15 & 84.02 & 27.67M & 171076M \\
\multirow{1}*{\{Warp-COCO,}&\textbf{DBase}  &30.94 &47.65 &32.66 & 10.82 & 27.76 & 43.32 & - & - & - & 32.29M & 174604M \\
\multirow{1}*{COCO\}}&\textbf{MBase}  & - & - & - & - & - & - & 38.94 & 58.67 & 78.39 & 23.51M & 143340M \\
&\textbf{MDBase}  &30.95 &47.57 &32.67 & 11.23 & 27.50 & 43.48 & 36.42 & 56.78 & 77.32 & 32.29M & 206641M \\
&\textbf{MatchDet-G}  &\textbf{43.62} &\textbf{61.66} &\textbf{47.32} & \textbf{19.71} & \textbf{41.90} & \textbf{56.74} & \textbf{63.18} & \textbf{77.86} & \textbf{89.02} & 38.54M & 248246M \\
&\textbf{MatchDet-P}  &{34.15} &{51.23} &{35.98} & {14.79} & {31.19} & {46.13} & {58.31} & {73.54} & {83.24} & 38.54M & 248246M \\
&\textbf{MatchDet-N}  &{32.28} &{49.28} &{34.01} & {12.81} & {29.40} & {44.38} & {54.11} & {69.06} & {79.82} & 38.54M & 257154M \\
\hline
&FCOS \cite{tian2019fcos}  &25.13 &39.56 &25.63 & 0.26 & 4.43 & 29.67 & - & - & - & 32.15M & 60709M \\
&LoFTR \cite{sun2021loftr}  & - & - & - & - & - & - & 10.75 & 31.48 & 50.02 & 27.67M & 69729M \\
\multirow{1}*{\{miniScanNet-F0,}&\textbf{DBase}  &24.99 &39.22 &25.50 & 0.24 & 4.39 & 29.52 & - & - & - & 32.15M & 60709M \\
\multirow{1}*{miniScanNet-F1\}}&\textbf{MBase}  & - & - & - & - & - & - & 5.87 & 16.34 & 33.66 & 23.51M & 50590M \\
&\textbf{MDBase}  &23.70 &38.05 &23.60 & 0.25 & 3.46 & 28.15 & 5.95 & 16.50 & 33.93 & 32.15M & 72016M \\
&\textbf{MatchDet-G}  &\textbf{40.69} &\textbf{58.12} &\textbf{41.38} & \textbf{1.88} & \textbf{25.76} & \textbf{45.03} & \textbf{18.43} & \textbf{39.14} & \textbf{57.11} & 38.39M & 100725M \\
&\textbf{MatchDet-P}  &28.29 &43.12 &29.34 & 0.82 & 6.54 & 33.03 & 14.39 & 35.18 & 53.24 & 38.39M & 100725M \\
&\textbf{MatchDet-N}  &27.33 &42.09 &27.58 & 0.56 & 5.47 & 32.29 & 12.47 & 33.09 & 51.32 & 38.39M & 104289M \\
\hline
\end{tabular}
\caption{MatchDet vs. other methods on different combinations of \{Target image, Reference image\} pairs on Full Warp-COCO and miniScanNet datasets. The \emph{Average Precision} on Target image, homography estimation AUC of the corner error in percentage, epochs=12, params and FLOPs are reported.
The DBase = $[{f_B},{f_F},{f_D}]$ and MBase = $[{f_B},{f_M}]$ are detector network and matcher network split from MDBase, respectively.
The MatchDet-G, MatchDet-P and MatchDet-N follow three different settings of \textbf{GTBoxR}, \textbf{PreBoxR} and \textbf{NoBoxR}, respectively.}
\label{table:sota}
\end{table*}

\renewcommand{\tabcolsep}{2.5pt}
\begin{table}[t!]
\centering
\small
\begin{tabular}{ l | c c c | c c c}
\hline
\multicolumn{1}{l|}{\multirow{2}*{Method}} &\multicolumn{3}{c|}{\emph{Ave. Precision}} & \multicolumn{3}{c}{Homo. est. AUC} \\
\cline{2-7} \multicolumn{1}{c|}{}  &AP & AP$_{50}$ & AP$_{75}$ &@3px &@5px &@10px\\
\hline
LoFTR  & - & - & - & 45.88 & 65.15 & 84.99 \\
\hdashline
FCOS  &37.32 &55.69 &40.37 & - & - & - \\
\textbf{DBase(FC)} & 37.19 &55.54 &39.96& - & - & - \\
\textbf{MBase} & - & - & -& 34.86 & 54.25 & 74.91 \\
\textbf{MDBase(FC)} &37.07 &55.38 &39.68 & 32.67 & 52.76 & 73.24 \\
\textbf{MatchDet-G(FC)} &\textbf{43.99} &\textbf{61.93} &\textbf{47.75} & \textbf{60.09} & \textbf{74.58} & \textbf{86.07} \\
\textbf{MatchDet-P(FC)} &{40.57} &{58.66} &{42.39} & {55.21} & {70.32} & {80.17} \\
\textbf{MatchDet-N(FC)} &{39.49} &{57.47} &{41.25} & {51.24} & {66.12} & {76.79} \\
\hdashline
Faster-RCNN & 35.90 & 56.28 & 39.05& - & - & - \\
\textbf{MatchDet-G(FR)} & \textbf{42.22} & \textbf{62.44} & \textbf{45.98}& \textbf{68.72} & \textbf{80.83} & \textbf{90.37} \\
\textbf{MatchDet-P(FR)} & 39.79 & 60.13 & 43.29& 65.13 & 79.02 & 87.56 \\
\textbf{MatchDet-N(FR)} & 37.85 & 58.72 & 41.55& 62.37 & 76.43 & 85.68 \\
\hdashline
AdaMixer & 40.82 & 59.72& 43.73& - & - & - \\
\textbf{MatchDet-G(AM)} & \textbf{48.19} & \textbf{66.57} & \textbf{52.35}& \textbf{68.88} & \textbf{80.96} & \textbf{90.96} \\
\textbf{MatchDet-P(AM)} & 45.01 & 64.35 & 49.23& 65.21 & 78.53 & 88.55 \\
\textbf{MatchDet-N(AM)} & 42.19 & 62.52 & 46.22& 62.74 & 76.69 & 86.88 \\
\hline
\end{tabular}
\caption{The results of our MatchDet with different detectors, on \{COCO,Warp-COCO\}.
The FC, FR, and AM represent FCOS, Faster-RCNN, and AdaMixer, respectively.}
\label{table:coco}
\end{table}

\subsection{Discussion}
\subsubsection{Task-collaborative vs. Task-individual}
Comparing to the current task-individual solutions, the advantages of our task-collaborative framework MatchDet are: (a) With the collaborative learning of image matching and object detection tasks, MatchDet is able to obtain mutual performance improvements. (b) Lower complexity with a shared backbone. (c) More convenient in practical applications with one single MatchDet model for two tasks.

\subsubsection{WSAM vs. WAM}
\label{sec:WSAM_WAM}
The WAM transforms the pair features of target and reference images into new feature space that are easy to measure similarity for matching, while the WSAM utilizes a spatial map to enhance the foregrounds of target image for detection.
Details are presented in APPENDIX.

\subsubsection{Weighted Attention vs. Masked Attention}
The proposed Weighted Attention has three main differences from Masked Attention \cite{cheng2022masked}, including input forms, weighting operation, and weights generation. These differences make Masked Attention can not be directly integrated by our MatchDet.
Masked Attention performs cross-attention between input features and learnable embedding, while Weighted Attention processes a pair of features.
Due to this difference in input forms, Weighted Attention needs to generate two corresponding weighted maps for feature pairs.
Besides, Mask2Former is applied in segmentation task with the supervision of ground-truth masks, which is easy to obtain good quality masks for Masked Attention.
While our MatchDet is proposed for detection task with ground-truth bounding boxes only, thus we propose a Weights Generator using the weakly-supervised segmentation technique to generate coarse masks for Weighted Attention.

\renewcommand{\tabcolsep}{2.5pt}
\begin{table}[t!]
\centering
\small
\begin{tabular}{ l | c c c | c c | c c c}
\hline
\multicolumn{1}{l|}{\multirow{2}*{Method}} & \multicolumn{3}{c|}{Modules} &\multicolumn{2}{c|}{\emph{Ave. Precision}} & \multicolumn{3}{c}{Homo. est. AUC} \\
\cline{2-9} \multicolumn{1}{c|}{}  &A. &S. &B.  &AP & AP$_{50}$ &@3px &@5px &@10px\\
\hline
\textbf{MDBase} &-&-&- &30.95 &47.57 & 36.42 & 56.78 & 77.32 \\
\hdashline
\textbf{MatchDet-G} &{\checkmark}&& &30.66 &47.25 & 60.12 & 74.92 & 87.01 \\
\textbf{MatchDet-G} &{\checkmark}&{\checkmark}& &\textbf{43.62} &\textbf{61.66} & {61.15} & {75.95} & {87.92} \\
\textbf{MatchDet-G} &{\checkmark}&{\checkmark}&{\checkmark} &\textbf{43.62} &\textbf{61.66} & \textbf{63.18} & \textbf{77.86} & \textbf{89.02} \\
\hdashline
\textbf{MatchDet-P} &{\checkmark}&& &30.69 &47.27 & 55.43 & 70.49 & 81.54 \\
\textbf{MatchDet-P} &{\checkmark}&{\checkmark}& &\textbf{34.15} &\textbf{51.23} & {56.42} & {71.48} & {82.35} \\
\textbf{MatchDet-P} &{\checkmark}&{\checkmark}&{\checkmark} &\textbf{34.15} &\textbf{51.23} & \textbf{58.31} & \textbf{73.54} & \textbf{83.24} \\
\hdashline
\textbf{MatchDet-N} &{\checkmark}&& &30.60 &47.21 & 51.67 & 66.52 & 78.20 \\
\textbf{MatchDet-N} &{\checkmark}&{\checkmark}& &\textbf{32.28} &\textbf{49.28} & 52.43 & 67.49 & 79.02 \\
\textbf{MatchDet-N} &{\checkmark}&{\checkmark}&{\checkmark} &\textbf{32.28} &\textbf{49.28} & \textbf{54.11} & \textbf{69.06} & \textbf{79.82} \\
\hline
\end{tabular}
\caption{The influence of WAM (A.), WSAM (S.) and Box Filter (B.) on \{Warp-COCO,COCO\} pairs dataset.}
\label{table:modules}
\end{table}

\begin{figure*}[t!]
\centering
\includegraphics[width=0.98\linewidth]{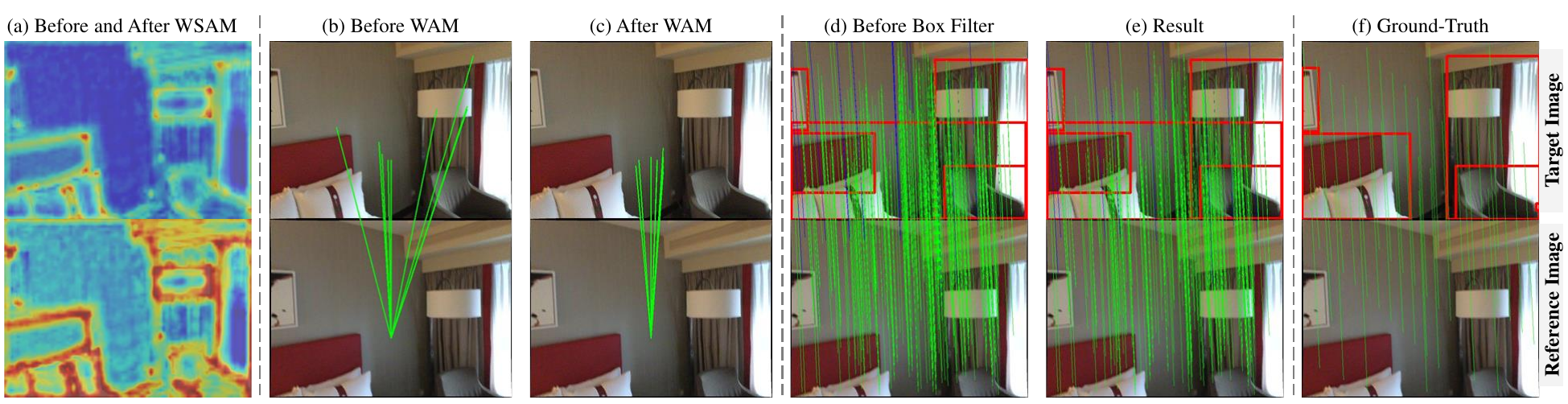}
\caption{The visualizations for WAM, WSAM, Box Filter and MatchDet results under \textbf{GTBoxR} setting from miniScanNet.
(a) - (e), are the results processed by the corresponding modules before and after, respectively.
(e) shows the predicted bounding boxes and matching results of MatchDet, where these matches are obtained after Box Filter.
(f) is Ground-Truth.}
\label{fig:visual_1}
\end{figure*}

\section{Experiments}
\subsection{Datasets}
\noindent\textbf{Warp-COCO} \
The Full Warp-COCO dataset consists of Warp-COCO and COCO \cite{lin2014microsoft}, i.e. every image in COCO has a corresponding transformed synthetic image to make a pair, which has ground-truth poses and boxes.

\noindent\textbf{miniScanNet} \
We selected 20 categories from ScanNet \cite{dai2017scannet} dataset (230M image pairs) with high-quality bounding boxes to make a new miniScanNet dataset (188K image pairs), which is 1000 times smaller than ScanNet.
The miniScanNet consists of real-world image pairs, which is more challenging than the synthetic Warp-COCO dataset.

\subsection{Evaluation Metrics and Implementation Details}
The \emph{Average Precision} (AP) and the Area Under the Cumulative curve (AUC) of the corner error, are used for object detection and image matching, respectively.
The used backbone is ResNet-50 \cite{he2016deep}.
The data augmentation is not applied for all methods.
The hyper-parameters are set as ${\alpha}_1$ = $1.0$, ${\alpha}_2$ = $1.0$, ${\beta}$ = $1.0$, ${\lambda}$ = $1.0$.

\renewcommand{\tabcolsep}{3pt}
\begin{table}[t!]
\centering
\small
\begin{tabular}{ l l | c c c | c c c}
\hline
\multicolumn{1}{l}{\multirow{2}*{$f_{WAM}$}} & \multicolumn{1}{l|}{\multirow{2}*{$f_{WSAM}$}} &\multicolumn{3}{c|}{\emph{Average Precision}} & \multicolumn{3}{c}{Homography est. AUC} \\
\cline{3-8}  &   &AP & AP$_{50}$ & AP$_{75}$ &@3px &@5px &@10px\\
\hline
C.A.&C.A. &31.01 &47.72 &32.91 & 45.84 & 66.13 & 82.89 \\
W.A.&W.A. &40.41 &58.43 &44.21 & 60.03 & 74.88 & 86.99 \\
W.S.A.&W.S.A. &42.13 &60.07 &46.00 & 58.46 & 73.40 & 86.08 \\
W.S.A.&W.A. &40.02 &58.00 &43.86 & 57.88 & 72.89 & 85.66 \\
W.A.&W.S.A. &\textbf{43.62} &\textbf{61.66} &\textbf{47.32} & \textbf{61.15} & \textbf{75.95} & \textbf{87.92} \\
\hline
\end{tabular}
\caption{The influence of Weighted Attention (W.A.) and Weighted Spatial Attention (W.S.A.), under \textbf{GTBoxR} setting on \{Warp-COCO,COCO\} pairs dataset.
The $f_{WAM}$ and $f_{WSAM}$ are inserted into Matcher and Detector respectively, i.e. $f_{WAM}$ and $f_{WSAM}$ directly affect the AUC and the AP respectively.
We give different combinations of Cross Attention (C.A.), W.A. and W.S.A.}
\label{table:WAM_WSAM}
\end{table}

\subsection{Comparison with Related Methods}
As shown in Tab.\ref{table:sota} and Tab.\ref{table:coco}, we conduct experiments on Full Warp-COCO and miniScanNet datasets with different combinations of \{Target image, Reference image\} pairs, including \{Warp-COCO, COCO\}, \{COCO, Warp-COCO\} and \{miniScanNet-F0, miniScanNet-F1\}.
The DBase and MBase are the two task-individual baselines for object detection and image matching, respectively.
The MDBase is the task-collaborative baseline for Match-and-Detection task, which obtains similar AP and AUC compared to DBase and MBase respectively.
Under three different settings of \textbf{GTBoxR}, \textbf{PreBoxR} and \textbf{NoBoxR}, our MatchDet outperforms these baselines on both AP and AUC performances, which demonstrates that the proposed approaches can achieve the collaborative learning between matching and detection tasks to obtain mutual performance improvements.
Under the \textbf{GTBoxR} setting, MatchDet obtains a larger performance improvement.
Even under the most challenging \textbf{NoBoxR} setting, our MatchDet still achieves competitive improvement.


\subsection{Ablation Study}
\noindent\textbf{WAM, WSAM and Box Filter.} \ \
The results in Tab.\ref{table:modules} demonstrate that the proposed three modules achieve consistent improvements under the settings of \textbf{GTBoxR}, \textbf{PreBoxR} and \textbf{NoBoxR}.
Specifically, the WSAM highlights the foreground regions of target image to boost the AP of detection task.
The WAM and Box Filter obtains high-quality matches to improve the AUC of image matching task.

\noindent\textbf{Weighted Attention and Weighted Spatial Attention.} \ \
The results in Tab.\ref{table:WAM_WSAM} show that:
(a) The best setting is using Weighted Attention for Matcher and Weighted Spatial Attention for Detector. These experimental results coincide with the theoretical analysis in AAPPENDIX.
(b) The proposed Weighted Attention and Weighted Spatial Attention, exploring the correlation between foreground regions of feature pairs, are able to obtain better performance than the traditional Cross Attention.

\subsection{Visualization Analysis}
Fig.\ref{fig:visual_1} shows the visualizations for WAM, WSAM, Box Filter and MatchDet results under \textbf{GTBoxR} setting.
Fig.\ref{fig:visual_1}(a) shows that the WSAM is able to highlight the foreground regions.
Comparing Fig.\ref{fig:visual_1}(c) to Fig.\ref{fig:visual_1}(b), the WAM finds the correspondences among surrounding regions to produce more discriminative feature representations.
Comparing Fig.\ref{fig:visual_1}(e) to Fig.\ref{fig:visual_1}(d), Box Filter reduces the background interference to obtain high-quality matches.


\section{Conclusion}
In this paper, we propose a collaborative framework called MatchDet for image matching and object detection to obtain mutual improvements.
To achieve the collaborative learning of the two tasks, three novel modules are proposed, including a Weighted Spatial Attention Module (WSAM) which highlights the foreground regions of target image for Detector, and Weighted Attention Module (WAM) and Box Filter which obtains high-quality matches for Matcher.
The experimental results on Warp-COCO and miniScanNet datasets show that our approaches are effective and achieve competitive improvements.

\bibliography{aaai24}

\section{APPENDIX}
\section{Related Work}
\noindent\textbf{Object Detection} \
The current object detection algorithms can be divided into anchor-based, anchor-free, and query-based methods.
The anchor-based two-stage \cite{fasterRCNN,maskRCNN} and multi-stage \cite{cascadeRCNN} methods utilize anchors to produce proposals for classification and box regression, and the anchor-based one-stage methods \cite{YOLOv3,retinaNet,freeAnchor,ATSS,guidedAnchoring} can directly classify and regress anchor boxes without relying on object proposal.
The anchor-free methods \cite{cornerNet,centerNet,objectsAsPoints,repPoints} implement object detection via detecting key-point or semantic-point, and another anchor-free methods \cite{denseBox,FSAF,foveaBox,tian2019fcos} obtain the detection results by densely classify each point on feature pyramids \cite{FPN}, and then predict the distances between the point and the four sides of box.
Recently, query-based detector, DETR \cite{carion2020end}, predicts a set of objects by attending queries to the feature map with the transformer decoder and achieves promising performance. Since the introduction of DETR, there have been several modifications and improvements \cite{zhu2020deformable,li2022dn,gao2022adamixer,zhang2022dino} proposed to overcome its limitations, such as slow training convergence and performance drops for small objects.
In this paper, we choose the classical anchor-free FCOS \cite{tian2019fcos} as the basic detector due to its good performance and simplicity.

\noindent\textbf{Image Matching} \
Image matching finds pixel-wise correspondences between pairs of images.
Early works focused on designing interest point detectors and descriptors \cite{moravec1981rover,harris1988combined,zhang1995robust,schmid1997local,lowe2004distinctive,dalal2005histograms,bay2006surf,rublee2011orb}.
Recently, learned interest point detectors and descriptors have been increasingly popular \cite{yi2016lift,detone2018superpoint,dusmanu2019d2,revaud2019r2d2,luo2019contextdesc,tyszkiewicz2020disk,wang2020learning}, typically outperforming the hand-crafted counterparts.
More recently, the interest point detector-free method, LoFTR \cite{sun2021loftr}, finds the correspondences by matching on dense and representative feature maps, without detecting keypoints. LoFTR \cite{sun2021loftr} and the follow-ups \cite{wang2022matchformer,chen2022aspanformer} match points distributed on dense grids rather than sparse locations, which boosts the robustness to impressive levels. MatchFormer \cite{wang2022matchformer} performs feature extraction and similarity learning through a transformer synchronously, which can provide matching-aware features in each stage of the hierarchical structure. To capture both global context and local details, ASpanFormer \cite{chen2022aspanformer} proposes a Transformer-based detector-free matcher, equipped with a hierarchical attention framework.
In this paper, we choose the classical LoFTR \cite{sun2021loftr} as the basic image matching method due to its good performance and simplicity.

\noindent\textbf{Transformer Attention}
Transformer is an attention-based architecture which is firstly introduced in natural language processing \cite{vaswani2017attention,devlin2018bert}.
Due to its powerful ability in learning representation, Transformer has been widely applied in vision tasks like image classification \cite{dosovitskiy2020image,wang2021pyramid,touvron2021training}, segmentation\cite{zheng2021rethinking,liang2020polytransform,xie2021segformer}, object detection\cite{zhang2020feature,carion2020end,zhu2020deformable} and image matching\cite{sun2021loftr,sarlin2020superglue}.
The Cross Attention used in LoFTR \cite{sun2021loftr} learns global context based on the full affinity matrix between feature pairs, but it may easily be distracted by background and causes the slow convergence of model.
To alleviate this problem, the Masked Attention was proposed in Mask2Former \cite{cheng2022masked}, which only made attention within the foreground regions of the predicted mask for query embedding.
Inspired by Masked Attention, we propose two novel transformer-based weighted attention and weighted spatial attention modules to strengthen the target object, by exploring the correlation between the foreground regions of feature pairs.

\section{Discussion}
\subsection{WSAM vs. WAM}
\label{sec:WSAM_WAM}
The WAM transforms the pair features of target and reference images into new feature space that are easy to measure similarity for matching, while the WSAM utilizes a spatial map to enhance the foregrounds of target image for detection.
Let's assume the features consist of foreground and background information.
Then we denote $C^t_3$ = $C^t_3[f]+C^t_3[b]$ and $C^r_3$ = $C^r_3[f]+C^r_3[b]$, where $[f]$ and $[b]$ represent the foreground and background information, respectively.
We assume the similarity scores are $\langle C^t_3[f], C^r_3[f] \rangle = v \in [0,1]$ and $\langle C^t_3[b], C^r_3[b] \rangle = u \in [0,1]$, and there is $v \approx u$ since the pair images are similar.
We next discuss the differences of feature processing by WAM and WSAM:

$\bullet$ {WAM} processing:
\begin{equation}
\begin{aligned}
\bar{C}^t_3[f] &\approx {C}^t_3[f] + v \cdot {C}^r_3[f] + (1-v) \cdot {C}^r_3[b], \\
\bar{C}^r_3[f] &\approx {C}^r_3[f] + v \cdot {C}^t_3[f] + (1-v) \cdot {C}^t_3[b]
\label{eq:wam_process}
\end{aligned}
\end{equation}
(a) For image matching task, the goal is to calculate the similarity between foreground regions $\bar{C}^t_3[f]$ and $\bar{C}^r_3[f]$.
According to Eq.\ref{eq:wam_process}, there is $\langle \bar{C}^t_3[f], \bar{C}^r_3[f] \rangle \approx v + v \cdot v + (1-v) \cdot u \approx 2v$, thus the similarity between foreground regions are enhanced with two times for Matcher.
(b) For object detection task, it aims to classify $\bar{C}^t_3[f]$, but the involved background information ${C}^r_3[b]$ may disturb the classification.
If $v<0.5$, the background disturbance of $(1-v) \cdot {C}^r_3[b]$ is even larger than the foreground enhancement $v \cdot {C}^r_3[f]$.

$\bullet$ {WSAM} processing:
\begin{equation}
\begin{aligned}
{{C}^t_3}'[f] &\approx {C}^t_3[f] + {C}^t_3[f] \cdot \langle {C}^t_3[f], \bar{C}^r_3[f] \rangle
\label{eq:wsam_process}
\end{aligned}
\end{equation}
where $w$ = $\langle {C}^t_3[f], \bar{C}^r_3[f] \rangle$ $\approx$ $v + v \cdot 1 + (1-v) \cdot (1-v)$ $\approx$ $1+v^2$. Thus the foreground ${C}^t_3[f]$ is strengthened with around $v^2$ times.

\subsection{MatchDet vs. Keypoint-base Tracking}
(1) We have discussed the differences between MatchDet and Tracker in the Introduction section. our MatchDet achieves a large improvement with 39.96\% in image matching task compared to Tracker.
(2) The framework of keypoint-base tracking \cite{nebehay2014consensus} is \{detect and describe keypoints $\rightarrow$ match keypoints $\rightarrow$ compute keypoints' displacement\}, which solves tracking problem by performing keypoints matching, without integrating object detection method. Differently, we propose a collaborative framework MatchDet for image matching and object detection to obtain mutual improvements.
In short, keypoint-base tracking is \{matching $\rightarrow$ tracking\}, while MatchDet is \{matching $\rightleftharpoons$ object detection\}.

\section{Experiments}
\subsection{Datasets}
\noindent\textbf{Warp-COCO} \
The Warp-COCO dataset consists of synthetic images generated from the large-scale detection benchmark COCO \cite{lin2014microsoft} by implementing perspective transformation.
Then the Full Warp-COCO dataset consists of Warp-COCO and COCO, i.e. every image in COCO has a corresponding transformed synthetic image to make a pair, which has ground-truth poses and bounding boxes.
Similar to the common practice \cite{ren2015faster,tian2019fcos} in detection task, we use the Full Warp-COCO \texttt{trainval35k} split (115K image pairs) for training, and \texttt{minival} split (5K image pair) for validation.
All images are resized to make the height 800 and the width less or equal to 1333.

\noindent\textbf{miniScanNet} \
The ScanNet \cite{dai2017scannet} dataset is an image matching benchmark, which has 230M image pairs with ground-truth poses and noisy bounding boxes.
We selected the samples of 20 categories with high-quality bounding box labels to make a new miniScanNet dataset, which is 1000 times smaller than ScanNet.
Specifically, 188K image pairs with overlap between [40\%, 80\%] are sampled for training, and 1500 testing pairs are used for evaluation.
Each image pairs \{miniScanNet-F0, miniScanNet-F1\} denote as the previous and the next frame images, respectively.
All images are resized to $640 \times 480$.
The miniScanNet consists of real-world image pairs, which is more challenging than the synthetic Warp-COCO dataset.

\subsection{Evaluation Metrics}
For object detection, consisting with FCOS \cite{tian2019fcos}, the most commonly metric \emph{Average Precision} (AP) is used, derived from precision and recall.
For image matching, following LoFTR \cite{sun2021loftr}, we report the area under the cumulative curve (AUC) of the corner error up to threshold values of 3, 5, and 10 pixels, respectively.
The corner error is computed between the images warped with the estimated $\hat{\mathcal{H}}$ and the ground-truth $\mathcal{H}$ as a correctness identifier as in \cite{sun2021loftr}.

\subsection{Implementation Details}
The used backbone is ResNet-50 \cite{he2016deep} with the initialized weights pre-trained on ImageNet \cite{deng2009imagenet}.
The data augmentation is not applied for all methods, and the total batch size is 16 pair images trained on 8 GPUs up-to 12 epoches.
The model is optimized end-to-end with stochastic gradient descent (SGD) with the initial learning rate of 0.01.
The learning rate is reduced by a factor of 10 after 8th epoch.
The weight decay and momentum are 0.0001 and 0.9, respectively.
The default hyper-parameters are set as $N_{WAM}$ = $2$, $N_{WSAM}$ = $1$, ${\alpha}_1$ = $1.0$, ${\alpha}_2$ = $1.0$, ${\beta}$ = $1.0$, ${\lambda}$ = $1.0$.
And other hyper-parameters are the same as FCOS \cite{tian2019fcos} and LoFTR \cite{sun2021loftr}.

\subsection{Ablation Study}
\noindent\textbf{Stacked Numbers of WAM and WSAM.} \ \
As illustrated in Tab.\ref{table:num_WAM_WSAM}, stacking more WAM is effective for image matching, while adding more WSAM is not much helpful for object detection.
The WAM transforms the pair features of target and reference images into new feature space, hence stacking more WAM is useful.
However, the WSAM utilizes a spatial map to enhance the foregrounds of target image, and stacking more WSAM only increases the scale of the spatial map, which is useless in further highlighting the target.

\renewcommand\thetable{5}
\renewcommand{\tabcolsep}{2.5pt}
\begin{table}[t]
\centering
\small
\begin{tabular}{ c c | c c c | c c c}
\hline
\multicolumn{1}{c}{\multirow{2}*{$N_{WAM}$}} & \multicolumn{1}{c|}{\multirow{2}*{$N_{WSAM}$}} &\multicolumn{3}{c|}{\emph{Average Precision}} & \multicolumn{3}{c}{Homography est. AUC} \\
\cline{3-8}  &   &AP & AP$_{50}$ & AP$_{75}$ &@3px &@5px &@10px\\
\hline
1&1 &43.44 &61.41 &47.10 & 57.93 & 72.44 & 84.99 \\
2&2 &43.39 &61.38 &47.00 & 60.89 & 75.32 & \textbf{87.96} \\
2&1 &\textbf{43.62} &\textbf{61.66} &\textbf{47.32} & \textbf{61.15} & \textbf{75.95} & {87.92} \\
\hline
\end{tabular}
\caption{The influence of stacked module numbers of WAM and WSAM (i.e. $N_{WAM}$ and $N_{WAM}$, respectively), under \textbf{GTBoxR} setting on \{Warp-COCO,COCO\} pairs dataset.}
\label{table:num_WAM_WSAM}
\vspace{-0.2cm}
\end{table}

\noindent\textbf{Influence of some hyper-parameters.} \ \
Influence of some hyper-parameters is shown in Tab.\ref{table:hyper}.

\renewcommand\thetable{6}
\renewcommand{\tabcolsep}{2.8pt}
\begin{table}[ht]
\centering
\small
\begin{tabular}{ c c c c | c c c | c c c}
\hline
\multicolumn{1}{c}{\multirow{2}*{${\alpha}_1$}} & \multicolumn{1}{c}{\multirow{2}*{${\alpha}_2$}} & \multicolumn{1}{c}{\multirow{2}*{${\beta}$}} & \multicolumn{1}{c|}{\multirow{2}*{${\lambda}$}} &\multicolumn{3}{c|}{\emph{Average Precision}} & \multicolumn{3}{c}{Homography est. AUC} \\
\cline{5-10}  & &&  &AP & AP$_{50}$ & AP$_{75}$ &@3px &@5px &@10px\\
\hline
0.5&0.5&0.5&0.5 &41.03 &59.05 &45.01 & \textbf{63.24} & \textbf{77.63} & \textbf{88.95} \\
0.5&0.5&0.5&1.0&42.59 &60.53 &46.21 & 61.92 & 76.13 & 88.00 \\
1.0&1.0&1.0&0.5&42.34 &60.22 &46.03 & {62.07} & {76.89} & {88.33} \\
1.0&1.0&1.0&1.0&\textbf{43.62} &\textbf{61.66} &\textbf{47.32} & {61.15} & {75.95} & {87.92} \\
\hline
\end{tabular}
\caption{The influence of some hyper-parameters, under \textbf{GTBoxR} setting on \{Warp-COCO,COCO\} pairs dataset.}
\label{table:hyper}
\vspace{-0.1cm}
\end{table}

\renewcommand\thefigure{7}
\begin{figure*}[ht]
\centering
\includegraphics[width=0.99\linewidth]{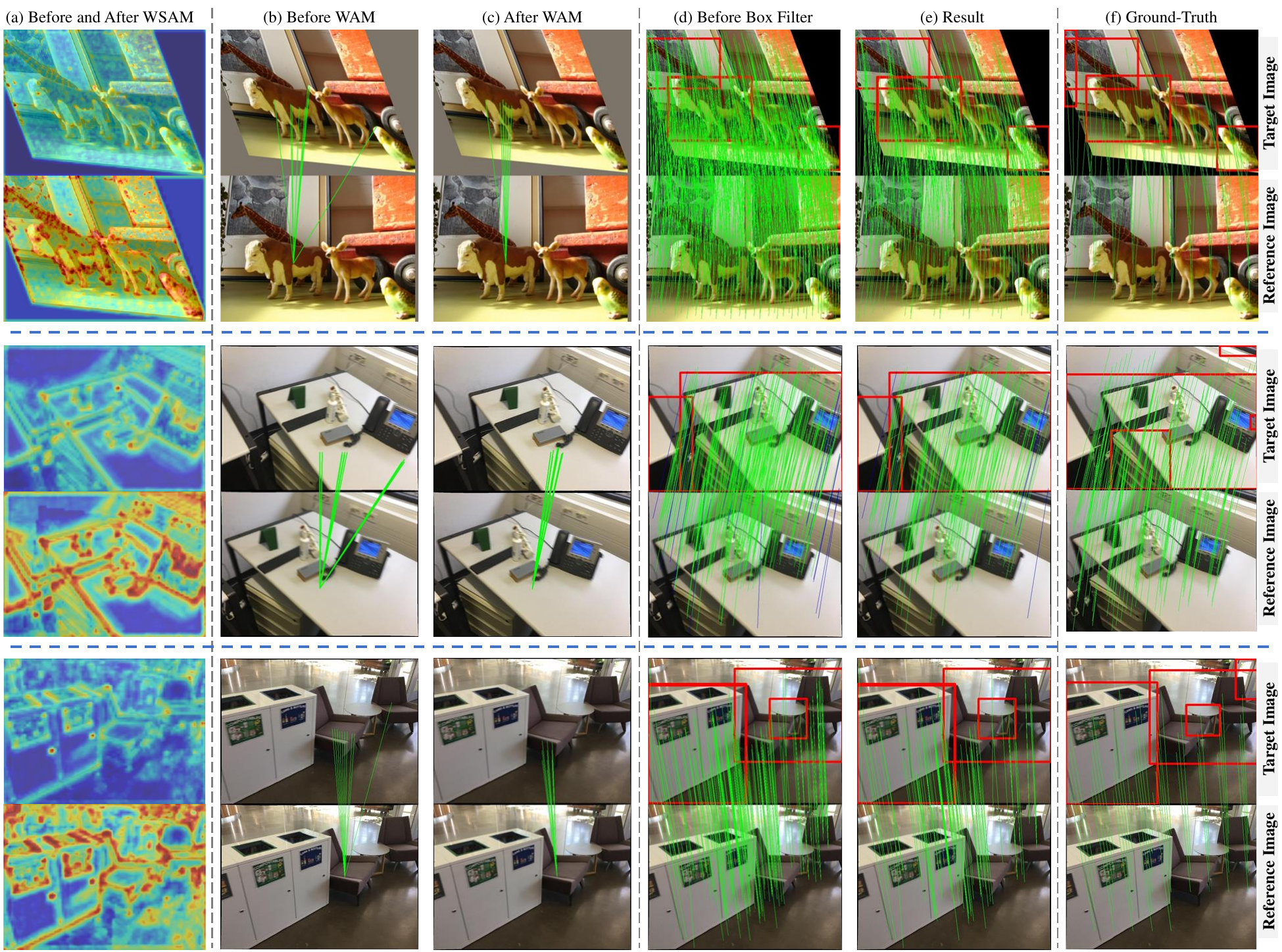}
\caption{The visualizations for WAM, WSAM, Box Filter and MatchDet results under \textbf{GTBoxR} setting.
(a) - (e), are the results processed by the corresponding modules before and after, respectively.
(e) shows the predicted bounding boxes and matching results of MatchDet, where these matches are obtained after Box Filter.
(f) is Ground-Truth.
The blue color lines indicate wrong matches.}
\label{fig:visual}
\end{figure*}

\renewcommand\thefigure{6}
\begin{figure}[H]
\centering
\includegraphics[width=0.99\linewidth]{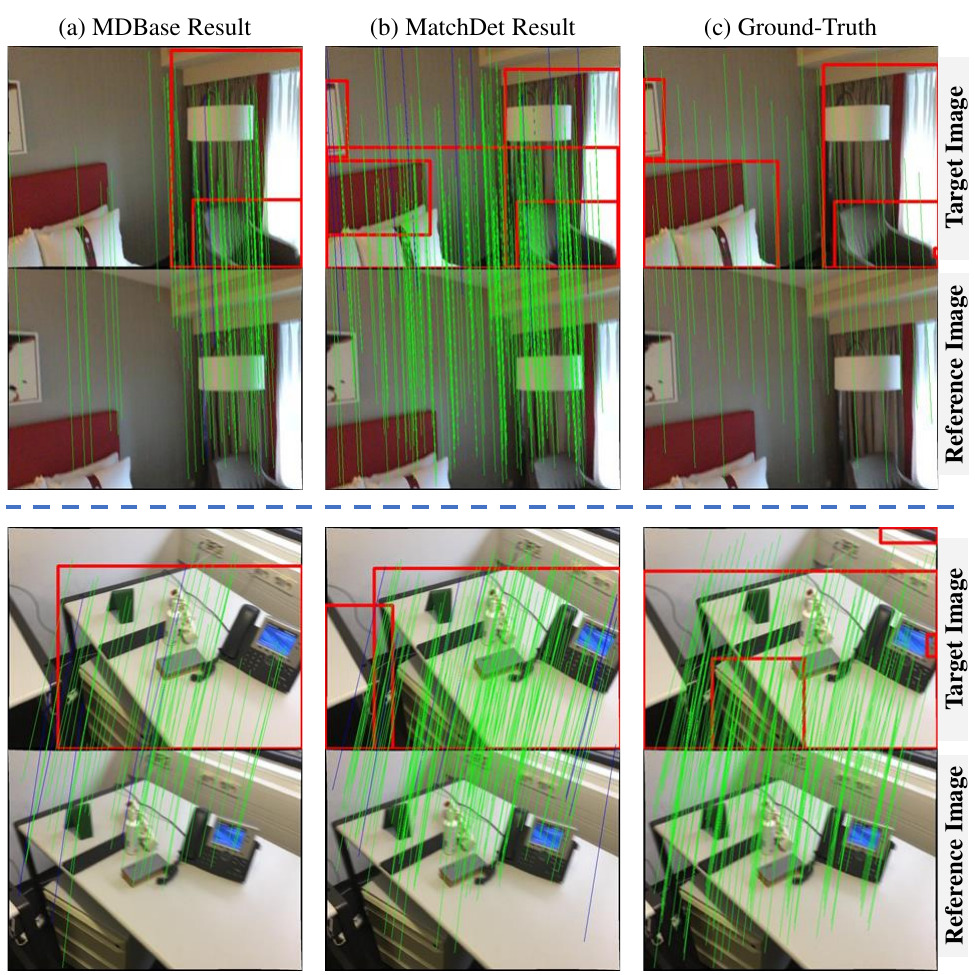}
\caption{The visualization comparisons between our MatchDet and MDBase.
Our MatchDet obtains better results of image matching and object detection.}
\label{fig:base_matchdet}
\end{figure}

\subsection{Visualization Analysis}
Fig.\ref{fig:base_matchdet} presents the visualization comparisons between our MatchDet and MDBase, which shows that our MatchDet obtains better results of image matching and object detection.
Fig.\ref{fig:visual} shows the visualizations for WAM, WSAM, Box Filter and MatchDet results.

\end{document}